\pdfoutput=1

\documentclass[11pt]{article}

\usepackage{EACL2023}

\usepackage{times}
\usepackage{latexsym}

\usepackage[T1]{fontenc}

\usepackage[utf8]{inputenc}

\usepackage{microtype}

\usepackage{inconsolata}

\usepackage{amsmath,amsfonts,bm}

\def\eqref#1{equation~\ref{#1}}

\def\1{\bm{1}}

\def\vz{{\bm{z}}}

\DeclareMathAlphabet{\mathsfit}{\encodingdefault}{\sfdefault}{m}{sl}
\SetMathAlphabet{\mathsfit}{bold}{\encodingdefault}{\sfdefault}{bx}{n}

\def\gF{{\mathcal{F}}}

\usepackage{hyperref}
\usepackage{url}
\usepackage{multirow,booktabs,makecell}
\usepackage{booktabs}
\usepackage{graphicx}
\usepackage{adjustbox}
\usepackage{color}	
\usepackage{comment}
\usepackage[labelformat=simple]{subcaption}

\usepackage[T1]{fontenc}
\usepackage{enumitem}
\usepackage{mathabx}
\usepackage{bm}
\usepackage{soul}
\usepackage{xspace}

\usepackage{cleveref}
\crefname{equation}{Eq.}{Eq.}
\crefname{section}{Section}{Sections}
\crefname{subsection}{Section}{Sections}
\crefname{subsubsection}{Section}{Sections}
\crefname{figure}{Figure}{Figures}
\crefname{table}{Table}{Tables}
\crefname{subfigure}{Figure}{Figures}
\crefname{algocf}{Algorithm}{Algorithms}

\newcommand{\imagine}[0]{\textsc{ImaginE}\xspace}
\newcommand{\imaginetext}[0]{\ensuremath{\imagine_{text}}\xspace}
\newcommand{\imagineimage}[0]{\ensuremath{\imagine_{image}}\xspace}
\newcommand{\imaginetextandimage}[0]{\ensuremath{\imagine_{text\&image}}\xspace}
\newcommand{\berttext}[0]{\ensuremath{\textsc{BERT}_{text}}\xspace}

\DeclareRobustCommand{\rev}[1]{{\sethlcolor{white}\hl{#1}}}
\soulregister{\ref}{7}
\soulregister{\cref}{7}
\soulregister{\cite}{7}
\soulregister{\citep}{7}
\soulregister{\citet}{7}
\soulregister{\imagine}{7}
\soulregister{\imaginetext}{7}
\soulregister{\imagineimage}{7}
\soulregister{\imaginetextandimage}{7}
\soulregister{\berttext}{7}

\title{\textsc{ImaginE}: An Imagination-Based Automatic Evaluation Metric \\for Natural Language Generation}

\author{
  Wanrong Zhu\textsuperscript{\P}, 
  Xin Eric Wang\textsuperscript{\S},
  An Yan\textsuperscript{\dag}, 
  Miguel Eckstein\textsuperscript{\P},
  William Yang Wang\textsuperscript{\P} 
  \\
  \textsuperscript{\P}UC Santa Barbara, 
  \textsuperscript{\S}UC Santa Cruz, 
  \textsuperscript{\dag}UC Santa Diego
  \\
  {\small \{wanrongzhu,william\}@cs.ucsb.edu, xwang366@ucsc.edu}\\
  {\small ayan@ucsd.edu, miguel.eckstein@psych.ucsb.edu}  \\
}

\begin{document}
\maketitle

\begin{abstract}
Automatic evaluations for natural language generation (NLG) conventionally rely on token-level or embedding-level comparisons with the text references. 
This is different from human language processing, for which visual imagination often improves comprehension. In this work, we propose \imagine, an imagination-based automatic evaluation metric for natural language generation. With the help of StableDiffusion~\citep{Rombach2022HighResolutionIS}, a state-of-the-art text-to-image generator, we automatically generate an image as the embodied imagination for the text snippet and compute the imagination similarity using contextual embeddings. 
Experiments spanning several text generation tasks demonstrate that adding machine-generated images with our \imagine displays great potential in introducing multi-modal information into NLG evaluation, and improves existing automatic metrics’ correlations with human similarity judgments in both reference-based and reference-free evaluation scenarios.
\end{abstract}
\section{Introduction}

A major challenge for natural language generation (NLG) is to design an automatic evaluation metric that can align well with human judgments. To this end, many approaches have been investigated. Metrics that base on matching mechanisms such as BLEU~\citep{BLEU}, METEOR~\citep{meteor}, CIDEr~\citep{cider}, have been widely adopted in the field. Edit-distance based metrics, such as CharacTER~\citep{Wang2016CharacTerTE}, WMD~\citep{Kusner2015FromWE}, SMD~\citep{Clark2019SentenceMS}, have also been explored. 
Recently, BERTScore~\citep{bert-score} and BLEURT~\citep{Sellam2020BLEURTLR} attempt to leverage BERT~\citep{Devlin2019BERTPO} to compare text embedding similarities, which correlates better with human judgments than previous methods.
These automatic evaluation metrics make use of textual information from various angles extensively.

But what happens in our minds when we read, comprehend, and evaluate text?
Research~\citep{Just2004ImageryIS,eviatar2006brain} has found that, unlike commonly designed automatic evaluation methods that compare the generated candidates with the references on the text domain only, humans, in contrast, leverage visual imagination and trigger neural activation in vision-related brain areas when reading text. Cognitive studies show that visual imagery improves comprehension during language processing~\citep{gambrell1986mental,joffe2007comprehension,sadoski2013imagery}. Inspired by this imagination-based multi-modal mechanism in human text comprehension,  we ask a critical research question: \textit{can machines create a visual picture of any underlying sentence, and use their imaginations to improve natural language understanding?} The advances of recent pre-trained vision-language models such as CLIP~\citep{clip} provide an excellent opportunity for us to utilize the learned image-text representations.
This enables us to explore the possibility of incorporating multi-modal information into NLG evaluation.

\begin{figure*}[t]
    \centering
    \includegraphics[width=\linewidth]{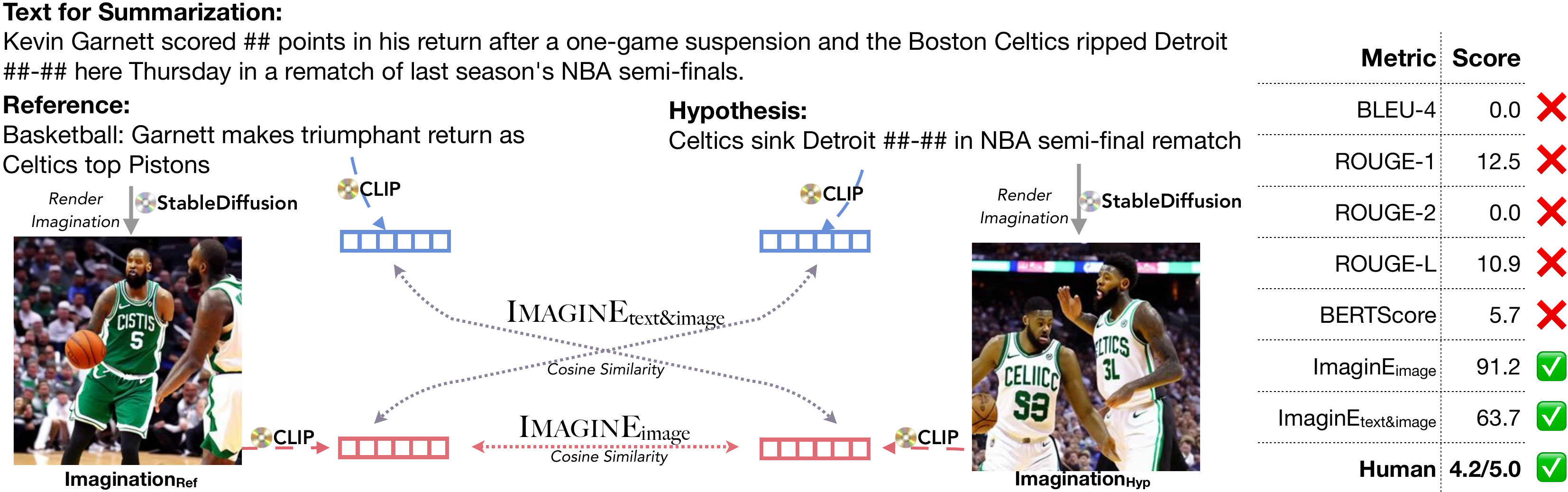}
\caption{An evaluation example on GigaWord for text summarization. \imagine visualizes machine imagination with StableDiffusion~\citep{Rombach2022HighResolutionIS} and extracts textual and visual representations with CLIP~\citep{clip}. While traditional evaluation metrics for natural language generation rely on  $n$-grams matching or textual embeddings comparison, 
\imagine incorporates machine-generated images into the evaluation process and enhances the understanding of the text snippet as a whole through the integration of multi-modal information.
\label{fig:example}}
\end{figure*}

In this work, we propose \imagine, an imagination-based automatic evaluation metric for text generation. Specifically, we first use the state-of-the-art text-to-image generator StableDiffusion~\citep{Rombach2022HighResolutionIS} to visualize machine imagination from sentences, which is to generate descriptive images for the candidate text and the references. 
\rev{Then we receive the \imagine~ scores by computing two sets of similarity scores with the pre-trained CLIP model~\citep{clip}: the visual similarity of the generated images, and the cross-modal similarity between the text and the generated image.}
\cref{fig:example} shows an example.

To understand the role the machine-generated images play in NLG evaluation, we conduct a series of experiments with \imagine on multiple NLG tasks and datasets, including machine translation, text summarization, and sentence completion for open-ended text generation, aiming to answer the following questions:
\begin{enumerate}[noitemsep,topsep=0pt,parsep=0pt,partopsep=0pt,leftmargin=*]
    \item \emph{How influential is \imagine in NLG evaluation in terms of correlations with human judgments? Can it provide additional reference information on top of existing metrics?}
    \item \emph{What are the applicable scenarios of introducing \imagine to NLG evaluation? When and why do machine-generated images help?}
    \item \emph{What are the potentials and limitations of introducing machine-generated images with \imagine to NLG evaluation?}
\end{enumerate}

Experimental results show that \imagine can serve as a complementary evaluation metric to text-based ones, and adding \imagine scores to existing metrics surprisingly improves most of the popular metrics' correlations with human performance on various text generation tasks. This holds for both reference-based evaluation and reference-free evaluation. 
We further conduct comprehensive quantitative analyses with case studies to verify its effectiveness.
Overall, \imagine displays great potential in introducing multi-modal information into NLG evaluation.

\section{Related Work}

\paragraph{Automatic Metrics for Natural Language Generation}
Common practices for NLG evaluation compare the generated hypothesis text with the annotated references. 
Metric performance is conventionally evaluated by its correlation with human judgments. Existing automatic evaluation metric calculations are mainly based on three mechanisms: $n$-grams overlap, edit distance, and embedding matching. 
BLEU~\citep{BLEU}, ROUGE-$n$~\citep{rouge}, METEOR~\citep{meteor} and CIDEr~\cite{cider} are a few widely used $n$-gram based metrics for text generation tasks. 
Another direction is based on edit distance~\citep{Toms2003AQM,Snover2006ASO,Panja2018ITERIT,Tillmann1997AcceleratedDB,Wang2016CharacTerTE}, where they calculate the edit distance between the two text snippets with different optimizations. 
Embedding-based metrics~\citep{Kusner2015FromWE,Rubner1998AMF,Clark2019SentenceMS,Lo2017MEANT2A,Lo2019YiSiA} evaluate text quality using word and sentence embeddings, and more recently, with the help of BERT~\citep{bert-score,Sellam2020BLEURTLR}.

\paragraph{Multi-Modal Automatic Metrics}
Aside from previous text-only metrics, some metrics utilize pre-trained multi-modal models and introduce visual features on top of text references for NLG evaluation. TIGEr~\citep{Jiang2019TIGErTG} computes the text-image grounding scores with pre-trained SCAN~\citep{Lee2018StackedCA}. ViLBERTScore-F~\citep{Lee2020ViLBERTScoreEI} relies on pre-trained ViLBERT~\citep{Lu2019ViLBERTPT} to extract image-conditioned embeddings for the text.
The CLIPScore~\citep{Hessel2021CLIPScoreAR} proposes a metric for image captioning by directly comparing images with captions using CLIP~\citep{clip}.
Our method differs in that we use visual picture generation as embodied imagination and apply our metric to various text-to-text generation tasks.

\paragraph{Mental Imagery}
The debate between pictorialists and propositionalists about how imagery information is stored in the human brain is still an open question in the neuroscience and psychology community~\citep{troscianko2013reading}. 
We follow the views from pictorialists that information can be stored in a depictive and pictorial format in addition to language-like forms~\citep{kosslyn2001neural,pearson2015heterogeneity}. In pictorialists’ model, mental imagery is constructed in the ``visual buffer'' either from the retinal image in seeing or from a long-term memory store of ``deep representations'' in the brain. Our image generation method is to mimic the generation of deep representations in machines, with the help of recent powerful text-to-image models. Inspired by empirical studies from cognitive science that visual imagination improves human text comprehension~\citep{gambrell1986mental,Sadoski1994ADC,nippold2003mental,Just2004ImageryIS,joffe2007comprehension,sadoski2013imagery}, we are interested in exploring if one can draw similar conclusions from automatic text evaluations by machines.

\begin{figure*}[t!]
    \centering
    \includegraphics[width=\linewidth]{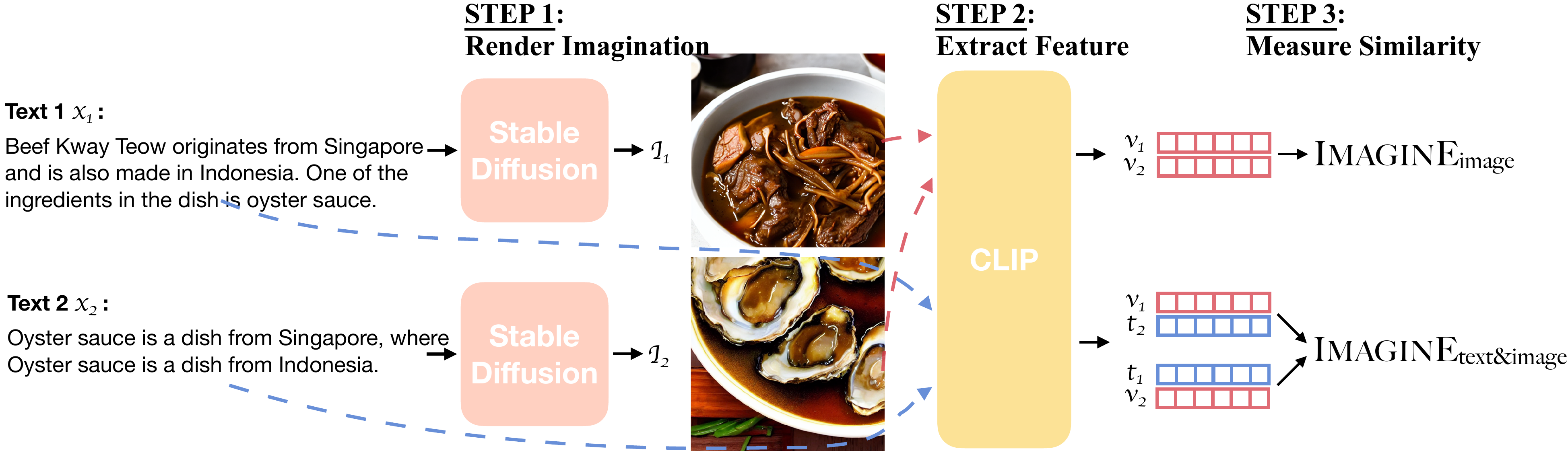}
\caption{Illustration of the computation process of the \imagine metric. 
Given the two pieces of text for comparison, $\bm{x}_1$ and $\bm{x}_2$, we render the machine imagination by generating two images $\bm{I}_1$ and $\bm{I}_2$ with the pre-trained StableDiffusion~\citep{Rombach2022HighResolutionIS}.
We extract features of the input text and corresponding generated images with CLIP~\citep{clip}.
We receive two variants of \imagine by computing the cosine similarity of the extracted features, in which \imagineimage measures mono-modal similarities on the visual side, while \imaginetextandimage conducts cross-modal matching.
\label{fig:overview}}
\end{figure*}

\section{\imagine}
\label{sec:method}

This section describes how our  \imagine~  metric evaluates the similarity between two pieces of text with the help of machine imagination. \ \cref{fig:overview} provides an overview of our method.

\subsection{Model Details}
\label{sec:pretrained_model}
\paragraph{CLIP} 
We use the cross-modal retrieval model, CLIP~\citep{clip}, for our evaluation purposes. CLIP jointly trains an image encoder and a text encoder to predict the correct pairing of image-text pairs with InfoNCE~\citep{Oord2018RepresentationLW} on 400M image-text pairs gathered from the web. 
We utilize the CLIP-\texttt{ViT-B/32} variant, which consists of a 12-layer, 8-head Transformer text encoder with a hidden size of 512, and a Vision Transformer (ViT)~\citep{Dosovitskiy2020AnII,Vaswani2017AttentionIA} image encoder adopting the BERT-\texttt{base} configuration and using a $32 \times 32$ input patch size. Both the text and image representations are normalized and projected into the multi-modal space before computing pairing likelihood through cosine similarity.

\paragraph{StableDiffusion} 
We perform text-to-image generation with StableDiffusion~\citep{Rombach2022HighResolutionIS}, which is a denoising diffusion probabilistic model~\citep{NEURIPS2020_4c5bcfec}. The model comprises three key components: a text encoder, a diffusion model, and an autoencoder. The text encoder, adopted from the frozen CLIP-\texttt{ViT-L/14}~\citep{clip}, is utilized to encode the input text into textual embeddings. The diffusion model, which leverages UNet~\citep{10.1007/978-3-319-24574-4_28} for noise estimation, is modified to attend to the input textual embeddings. We conduct experiments with StableDiffusion-\texttt{v1-1}, which was trained with LAION~\citep{schuhmann2022laionb}, using $256 \times 256$ images for pre-training, followed by $512 \times 512$ images for fine-tuning.

\subsection{\imagine Similarity Score}
\label{sec:imagine_score}

In our proposed approach, as depicted in Figure~\ref{fig:overview}, the computation of \imagine~ consists of three sequential steps. Firstly, the StableDiffusion model~\citep{Rombach2022HighResolutionIS} is utilized to generate descriptive images, referred to as machine imagination, from the two text snippets being compared. Secondly, both the text snippets and the generated images are encoded using the CLIP model~\citep{clip}. Finally, \imagine~ is calculated by computing the cosine similarities of the resulting text and visual features, both in a mono-modal and cross-modal manner.

\paragraph{Step 1: Render Imagination}
For each image, StableDiffusion randomly initializes a latent matrix $\bm{H}$ from the standard normal distribution and uses the encoder of the pre-trained autoencoder to encode $\bm{H}$ into the lower-resolution latent map $\vz_T$ ($T$ is the total inference steps).
At each step $t$, the diffusion model estimates the noise, $\epsilon$, and subtracts it from $\vz_t$.
The decoder of the pretrained autoencoder takes the final noise-free latent map $\vz$ and generates the image prediction $\bm{I}$ of size $512 \times 512$.

\paragraph{Step 2: Extract Feature}
In the previous step, we generate the corresponding images $\bm{I}_1$ and $\bm{I}_2$ for the pair of text $\bm{x}_1$ and $\bm{x}_2$ for comparison  with the text-to-image synthesis backbone.
Then we pass the machine-generated images $\bm{I}_1$ and $\bm{I}_2$ and the input text $\bm{x}_1$ and $\bm{x}_2$ through corresponding CLIP encoders to receive the visual representations $\bm{v}_1$,  $\bm{v}_2$, and the textual representation $\bm{t}_1$,  $\bm{t}_2$.

\paragraph{Step 3: Measure Similarity}
With $\mathrm{sim}(\cdot,\cdot)$ denoting the process of first normalizing the two vectors, then computing their cosine similarity, we compute two types of similarity scores for \imagine with the extracted textual and visual features: 

(1) \imagineimage computes the visual representation similarity between $\bm{v}_1$ and $\bm{v}_2$:
\begin{equation}
    \imagineimage = \gF \left( 
    \mathrm{sim}(\bm{v}_1, \bm{v}_2)
    \right)
\end{equation}

(2) \imaginetextandimage~ ($\imagine_{t\&i}$) takes both the text and the generated image into consideration, and conducts cross-modal comparisons between ($\bm{t}_1$, $\bm{v}_2$), as well as ($\bm{t}_2$, $\bm{v}_1$):
\begin{equation}
\imagine_{t\&i}  =  \gF \left( 
    \frac{\mathrm{sim}(\bm{t}_1,\bm{v}_2) + \mathrm{sim}(\bm{t}_2,\bm{v}_1)}{2}
    \right)
\end{equation}

The cosine similarity between the text and image representations theoretically has a range of $\left[ -1, 1\right]$. However, in practice, the \imagine similarity scores tend to cluster within a more narrow interval $\left[ l, h\right]$. 
Following \citet{Hessel2021CLIPScoreAR}, we use a linear function $\gF$ to stretch the similarity score distribution to the range of $\left[ 0, 1\right]$, which is also the score range for most of the automatic metrics covered in this study. \cref{eq:rescale} shows how we re-scale the similarity score $s$ into $s'$. Appendix Figure~\ref{fig:score_distribution} plots the two \imagine variants' distributions before and after rescaling.
\begin{equation}
\begin{split}
   \vspace{-1em}
    s' &= \frac{s - l}{h - l},\\ 
    \left[ l, h\right] &= 
    \left\{
        \begin{aligned}
        \left[0.1, 1.0\right] & , \; \text{for} \; \imagineimage, \\
        \left[0.1, 0.4 \right]  & , \; \text{for} \; \imaginetextandimage.
        \end{aligned}
    \right.
\label{eq:rescale}
\end{split}
\end{equation}

\subsection{Integration with Existing Metrics}
\label{sec:usage}

The \imagine similarity scores can serve as standalone automatic metrics. Additionally, \imagine can be incorporated as an extension to existing metrics, as it offers multimodal references and addresses the limitations of current text-only evaluations that only compare tokens or text embeddings. This mimics the human process of comprehending text, where both text and visual imagination are utilized. The integration of \imagine with other automatic metrics is straightforward, achieved by summing the \imagine similarity score with the other automatic metric's score for each example:
\begin{equation}
    metric\_score' \mathrel{{+}{=}} \imagine_{similarity\_score}
\end{equation}

\begin{table*}[t]\centering
\begin{adjustbox}{width=0.7\linewidth,center}
\begin{tabular}{l|rrr|rrr}\toprule
\multirow{2}{*}{\textbf{Metric}} &\multicolumn{3}{c|}{\textbf{IWSLT'14}} &\multicolumn{3}{c}{\textbf{WMT'19}} \\
\cmidrule(lr){2-4}\cmidrule(lr){5-7}
&\textbf{Original} & $+\text{IE}_{image}$ &$+\text{IE}_{text\&image}$ &\textbf{Original} &$+\text{IE}_{image}$ &$+\text{IE}_{text\&image}$ \\\cmidrule{1-7}
BLEU-1 &21.47 &21.38$\pm$1.53 &\textbf{21.86}$\pm$0.82 &13.74 &14.71$\pm$1.19 &\textbf{16.40}$\pm$0.73 \\
BLEU-2 &20.82 &21.17$\pm$1.45 &\textbf{21.53}$\pm$0.68 &12.50 &12.93$\pm$1.13 &\textbf{15.11}$\pm$0.64 \\
BLEU-3 &19.17 &19.88$\pm$1.39 &\textbf{20.31}$\pm$0.62 &11.31 &12.07$\pm$1.09 &\textbf{13.90}$\pm$0.58 \\
BLEU-4 &17.60 &18.57$\pm$1.36 &\textbf{19.08}$\pm$0.60 &9.10 &9.15$\pm$1.06 &\textbf{11.84}$\pm$0.54 \\
METEOR &20.60 &21.44$\pm$1.54 &\textbf{21.30}$\pm$0.99 &13.47 &14.77$\pm$1.33 &\textbf{16.80}$\pm$0.91 \\
ROUGE &20.55 &20.69$\pm$1.54 &\textbf{21.26}$\pm$0.80 &11.40 &11.58$\pm$1.16 &\textbf{14.34}$\pm$0.68 \\
CIDEr &21.98 &22.12$\pm$0.24 &\textbf{22.25}$\pm$0.07 &11.82 &11.86$\pm$0.18 &\textbf{12.05}$\pm$0.07 \\
BERTScore &23.95 &24.02$\pm$1.41 &\textbf{24.09}$\pm$0.65 &17.01 &17.08$\pm$1.22 &\textbf{18.88}$\pm$0.78 \\
BLEURT &22.93 &22.99$\pm$0.64 &\textbf{23.40}$\pm$0.41 &18.81 &19.36$\pm$0.82 &\textbf{19.59}$\pm$0.37 \\
\bottomrule
\end{tabular}
\end{adjustbox}
\caption{
The effect of applying our \imagine~ similarities on automatic metrics for machine translation, reflected in the Pearson correlation with human judgments. 
The image generation process is conducted over five different random seeds for each piece of text. We report the mean and standard deviation of the repeated runs.
IE: \imagine.
}
\label{tab:machine_translation_pearson}
\end{table*}

\section{Experimental Setup}
\subsection{Tasks, Datasets, and Models}
We evaluate our approach on three popular natural language generation tasks: machine translation, abstractive text summarization, and open-ended text generation.

\paragraph{Machine Translation} We use Fairseq~\citep{ott2019fairseq} to generate English translation from German on IWSLT'14~\citep{Bell201411thIW} and WMT'19~\citep{Barrault2019FindingsOT} datasets. 

\paragraph{Abstractive Text Summarization} We use the implementation of ~\citet{Li2017DeepRG} to generate summarization on DUC2004\footnote{\href{https://duc.nist.gov/duc2004/}{https://duc.nist.gov/duc2004/}} and use ProphetNet~\citep{Yan2020ProphetNetPF} for generation on Gigaword.\footnote{\href{https://catalog.ldc.upenn.edu/LDC2011T07}{https://catalog.ldc.upenn.edu/LDC2011T07}} Both datasets are built upon news articles.

\paragraph{Open-ended Text Generation} 
We perform experiments on the ActivityNet~\citep{Heilbron2015ActivityNetAL} subset of HellaSwag~\citep{zellers-etal-2019-hellaswag}, which is a benchmark for commonsense natural language inference that ask the model to predict the most likely follow-up among several choices given a specific context. The dataset is derived from ActivityNet video captions and we use it for the task of sentence completion, where the model is given a context and asked to complete the sentence. The predicted sentence endings generated by StoryEndGen~\citep{Guan2019StoryEG} and GPT-2~\citep{Radford2019LanguageMA} are collected and used in the following evaluation. 

\subsection{Automatic Metrics}

\paragraph{Machine Translation \& Summarization}
In the evaluation of machine translation and text summarization tasks, it is a common practice to compare the predicted text with the reference. Adhering to previous studies, we present results using reference-based metrics. For machine translation, we present scores using BLEU-$n$ ($n$=1,2,3,4)~\citep{BLEU}, METEOR\citep{meteor}, and CIDEr~\cite{cider}. Meanwhile, for text summarization, we present ROUGE-$n$ ($n$=1,2)~\citep{rouge} precision scores. Additionally, we report the scores of ROUGE-L~\citep{rouge}, BERTScore~\citep{bert-score}, and BLEURT~\citep{Sellam2020BLEURTLR} for both tasks.

\begin{table*}[t]\centering
\begin{adjustbox}{width=0.7\linewidth,center}
\begin{tabular}{l|rrr|rrr}\toprule
\multirow{2}{*}{\textbf{Metric}} &\multicolumn{3}{c|}{\textbf{DUC2004}} &\multicolumn{3}{c}{\textbf{GigaWord}} \\
\cmidrule(lr){2-4}\cmidrule(lr){5-7}
&\textbf{Original} & $+\text{IE}_{image}$ &$+\text{IE}_{text\&image}$ &\textbf{Original} &$+\text{IE}_{image}$ &$+\text{IE}_{text\&image}$ \\\cmidrule{1-7}
ROUGE-1 &13.66 &\textbf{16.77}$\pm$1.31 &13.45$\pm$0.80 &12.90 &\textbf{17.52}$\pm$0.73 &16.78$\pm$0.66 \\
ROUGE-2 &9.74 &\textbf{15.71}$\pm$1.65 &11.19$\pm$1.08 &7.75 &\textbf{14.26}$\pm$0.83 &13.33$\pm$0.77 \\
ROUGE-L &13.14 &\textbf{16.35}$\pm$1.47 &13.17$\pm$0.95 &14.31 &\textbf{17.44}$\pm$0.77 &16.78$\pm$0.70 \\
BERTScore &19.44 &\textbf{20.60}$\pm$1.29 &20.26$\pm$0.78 &19.59 &\textbf{20.47}$\pm$0.64 &20.10$\pm$0.57 \\
BLEURT &23.59 &\textbf{25.20}$\pm$0.72 &24.46$\pm$0.42 &20.23 &\textbf{21.08}$\pm$0.39 &20.74$\pm$0.35 \\
\bottomrule
\end{tabular}
\end{adjustbox}
\vspace{-5px}
\caption{
The effect of applying our \imagine~ similarities on automatic metrics for text summarization, reflected in the Pearson correlation with human judgments. 
The image generation process is conducted over five different random seeds for each piece of text. We report the mean and standard deviation of the repeated runs.
IE: \imagine.
}
\label{tab:summarization_pearson}
\end{table*}
\begin{table*}[t]\centering
\begin{adjustbox}{width=0.7\linewidth,center}
\begin{tabular}{l|rrr|rrr}\toprule
\multirow{2}{*}{\textbf{Metric}} &\multicolumn{3}{c|}{\textbf{Reference-based}} &\multicolumn{3}{c}{\textbf{Reference-free}} \\\cmidrule{2-7}
\cmidrule(lr){2-4}\cmidrule(lr){5-7}
&\textbf{Original} & $+\text{IE}_{image}$ &$+\text{IE}_{text\&image}$ &\textbf{Original} &$+\text{IE}_{image}$ &$+\text{IE}_{text\&image}$ \\\cmidrule{1-7}
div-2 &27.21 &28.01$\pm$0.49 &\textbf{28.08}$\pm$0.34 &27.21 &26.51$\pm$0.42 &\textbf{27.29}$\pm$0.58 \\
div-3 &26.80 &27.67$\pm$0.49 &\textbf{27.78}$\pm$0.35 &26.80 &26.17$\pm$0.43 &\textbf{26.98}$\pm$0.59 \\
div-4 &26.20 &27.14$\pm$0.48 &\textbf{27.28}$\pm$0.36 &26.20 &25.71$\pm$0.44 &\textbf{26.55}$\pm$0.60 \\
diversity &27.40 &28.19$\pm$0.41 &\textbf{28.23}$\pm$0.30 &27.40 &26.89$\pm$0.36 &\textbf{27.55}$\pm$0.50 \\
distinct-2 &26.72 &27.76$\pm$0.56 &\textbf{27.90}$\pm$0.40 &26.72 &25.54$\pm$0.48 &\textbf{26.49}$\pm$0.66 \\
BERTScore &23.47 &\textbf{25.92}$\pm$0.50 &25.43$\pm$0.36 &25.10 &23.47$\pm$0.56 &\textbf{25.26}$\pm$0.78 \\
BLEURT &19.99 &\textbf{22.47}$\pm$0.83 &21.55$\pm$0.72 &18.70 &19.67$\pm$0.88 &\textbf{20.56}$\pm$1.25 \\
\bottomrule
\end{tabular}
\end{adjustbox}
\vspace{-5px}
\caption{
The effect of applying our \imagine~ similarities on ActivityNet for open-ended text generation, reflected in the Pearson correlation with human judgments. 
In the ``Reference-based'' setting, we compare the predictions with the references, while in the ``Reference-free'' setting, we compare the predictions with the input contexts.
The image generation process is conducted over five different random seeds for each piece of text. We report the mean and standard deviation of the repeated runs.
IE: \imagine.
}
\vspace{-10px}
\label{tab:sentence_completion_pearson}
\end{table*}

\paragraph{Open-ended Text Generation}
In the context of open-ended text generation, where the number of possible answers for a given scenario can be inexhaustible, evaluating the quality of generated text through a comparison with a fixed set of references is challenging. To address this issue, previous studies have proposed to utilize reference-free metrics to evaluate the quality of the generated text. In this work, we experiment with the following reference-free metrics which assess model degeneration:
(1) div-$n$ = $\frac{|\text{unique }n\text{-grams}|}{|\text{total }n\text{-grams}|}$ measures sequence level repetition by computing the portion of duplicate $n$-grams ($n$=2,3,4) ~\citep{DBLP:conf/iclr/WelleckKRDCW20}. 
(2) diversity = $\prod_{n\text{=2}}^{4}\text{rep-}n$ measures the diversity of $n$-grams~\citep{su2022contrastive}, and assesses the model degeneration.
(3) distinct-$n$ = $\frac{|\text{unique }n\text{-grams}|}{|\text{length of text}|}$ measures the portion of distinct $n$-grams (here $n$=2)  in the text~\citep{Li2016ADO}. 
In addition, we report results on BERTScore~\citep{bert-score} and BLEURT~\citep{Sellam2020BLEURTLR} for comparison of contextual similarity.

\subsection{Human Evaluation}

We invite Amazon Mechanical Turk\footnote{\href{https://www.mturk.com/}{https://www.mturk.com/}} annotators to evaluate the quality of the generated text.
Due to cost constraints, when conducting human evaluation, we randomly sample 1,000 test examples for each dataset, except for DUC2004 which has 500 examples in the test set. Each example is evaluated by three human judges using a 5-point Likert scale, which assessed the fluency, grammar correctness, and factual consistency of the generated text with the reference text. The overall human assessment score is calculated as the mean of the scores obtained from the three aspects. We compute the Pearson correlation~\citep{freedman2007statistics} between the human scores and the scores obtained from the automatic metrics, and the results are reported as a multiple of 100 for clarity.

\section{Results and Analysis}

\subsection{Main Results}
\label{sec:results}

\paragraph{Machine Translation}
\label{sec:translation}


Table~\ref{tab:machine_translation_pearson} presents the results of the system-level Pearson correlation with human judges when extending the \imagine similarity metric to various existing automatic natural language generation (NLG) metrics on the IWSLT'14 and WMT'19 German-to-English datasets. The results demonstrate that the addition of both \imagineimage and \imaginetextandimage improves the Pearson correlation for all metrics listed. Among the two variants, the mean of \imaginetextandimage consistently performs better on both datasets. It is observed that there is a more substantial variance in \imagineimage, which is attributed to the difference in the images generated by the StableDiffusion model~\citep{Rombach2022HighResolutionIS} due to varying random seed and initialization values. As a result, \imagineimage, which compares two machine-generated images, has a higher standard deviation compared to \imaginetextandimage.

\paragraph{Abstractive Text Summarization}
\label{sec:summarization}


The results in Table~\ref{tab:summarization_pearson} demonstrate the system-level Pearson correlation with human judges when incorporating our \imagine similarity into existing automatic NLG metrics on the DUC2004 and Gigaword datasets.
In alignment with the observations made in the machine translation task, the addition of both \imagineimage and \imaginetextandimage results in an improvement in Pearson correlation across all metrics.
On the two summarization datasets, we notice that the correlation after incorporating \imagineimage exhibits higher mean values along with larger variances compared to the correlation with \imaginetextandimage.

\paragraph{Open-ended Text Generation}
\label{sec:open-ended}

For the sentence completion task, we conduct evaluations in two setups. In the reference-based evaluation, we compare the predicted sentence ending with the ground-truth ending provided in the dataset. 
In reference-free evaluation, we compare the predicted sentence ending with the input context. This setup is designed to assess the coherence of the prediction with the input context, as it is hypothesized that a high-quality prediction for open-ended text generation should be consistent with the input context.

The results of extending our \imagine similarity metric to existing automatic NLG metrics for the sentence completion task on the ActivityNet dataset are shown in \cref{tab:sentence_completion_pearson}.
In the reference-based setting, both \imagine variants demonstrate improvement over the listed metrics and exhibit comparable performances. In the reference-free setting, the introduction of \imaginetextandimage continues to enhance the Pearson correlation, while the implementation of \imagineimage results in a decrease in correlation. 
One possible reason for the decline in correlation when \imagineimage is used in the reference-free setting of the sentence completion task on ActivityNet (which is comprised of video captions) is that, despite the requirement for the predicted continuation to be coherent with the given context, the visual representation of the context and continued text may differ greatly in this scenario (e.g., due to a plot twist in the video). Consequently, direct comparison of images through \imagineimage may result in a decrease in correlation.
However, the inherent coherence between the input text and the continued text may be captured through cross-modal comparison, which may explain why \imaginetextandimage still improves the correlation for the listed metrics.

\subsection{Performance Analysis}

\label{sec:discussion}

\paragraph{Why is ImaginE helpful?}
As shown in~\cref{tab:machine_translation_pearson,tab:summarization_pearson,tab:sentence_completion_pearson}, the incorporation of certain variants of \imagine improves the correlation between the reference-based and reference-free metrics and human scores in the majority of cases. This indicates the usefulness of extending text-only metrics with multi-modal knowledge. 
However, how do these machine imaginations actually help text understanding and evaluation? In this section, we further explore how and why \imagine works. We first provide a case study to show the uniqueness of \imagine over text-based metrics, then systematically analyze the effectiveness of our method from different perspectives.

\begin{figure}[t!]
    \centering
    \includegraphics[width=\linewidth]{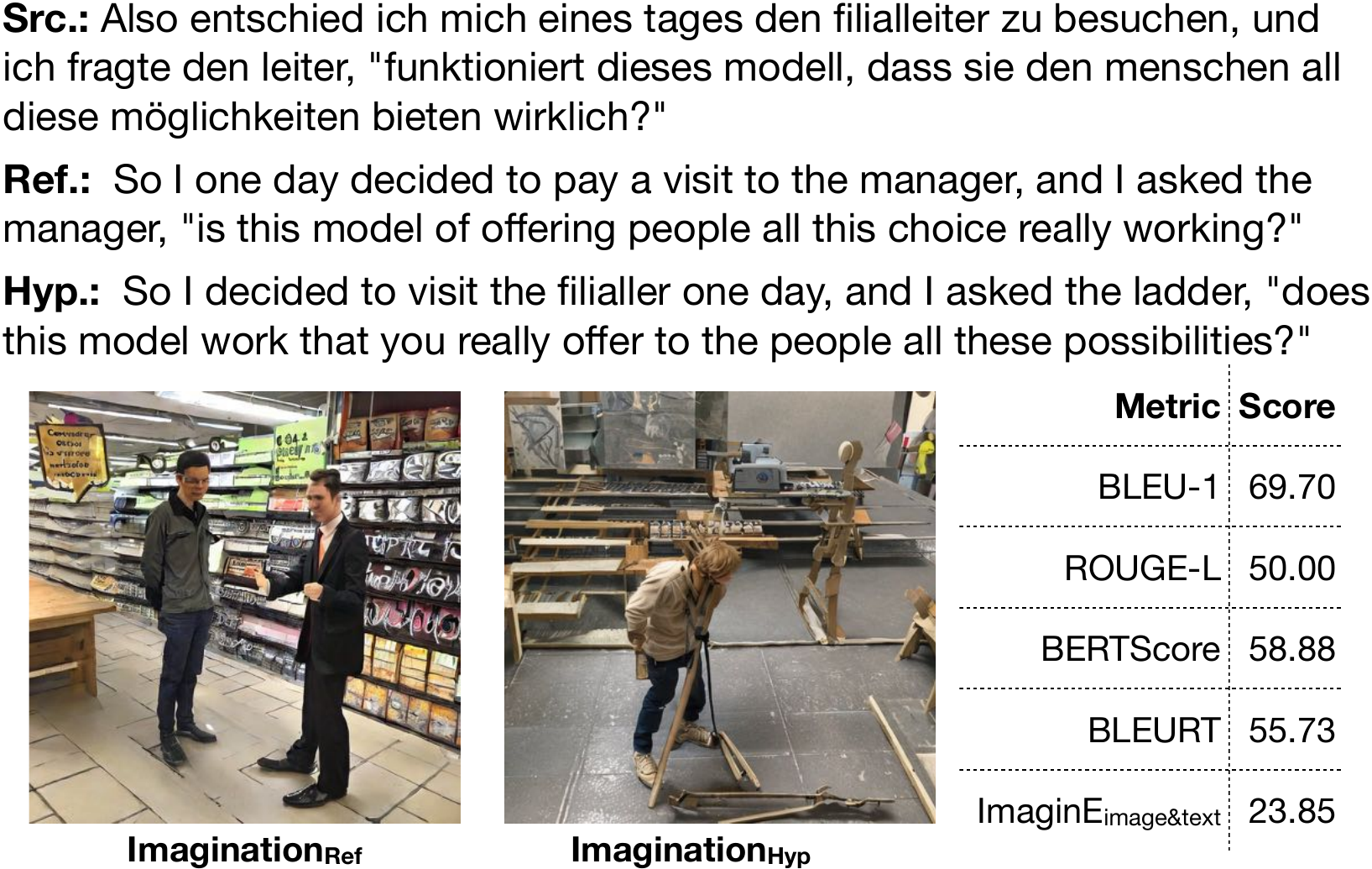}
\vspace{-10px}
\caption{
A case study on IWSLT'14 German-to-English translation with images rendered by StableDiffusion-\texttt{v2-1}. Src.: input source text. Ref.: reference text. Hyp.: generated hypothesis text.
\label{fig:word_choice}
}
\end{figure}

\paragraph{Case Study}
Figure~\ref{fig:word_choice} shows an example in which \imagine effectively detects the dissimilarity in keywords between two text snippets.
Despite the similarity in sentence structure between the reference and hypothesis, the crucial distinction lies in the inclusion of the terms ``manager'' and ``ladder''. While traditional automatic metrics that rely on $n$-grams matching (BLEU, ROUGE) or textual embedding comparison (BERTScore, BLEURT) may exhibit high scores, the quality of the generated text remains questionable. In contrast, \imagine generates distinctive images and exhibits a relatively low cross-modal similarity score, which aligns with human perception.

\begin{table}[t!]
\begin{adjustbox}{width=\linewidth,center}
\begin{tabular}{ c | r | r r r r}
\cmidrule[\heavyrulewidth]{1-6}
\textbf{Metric} & \textbf{Original} &	+$\textsc{IE}_{i(\text{dVAE})}$	&	+$\textsc{IE}_{i(\text{BigGAN})}$   &	+$\textsc{IE}_{i(\text{VQ-GAN})}$ &	+$\textsc{IE}_{i(\text{SD})}$\\ \cmidrule[\heavyrulewidth]{1-6}
ROUGE-1	&	13.7	&	15.9	$\pm$	0.9	&	15.7	$\pm$	1.0	&	15.9	$\pm$	0.8	 &  \textbf{16.8} $\pm$ 1.3\\
ROUGE-2	&	9.7	&	14.9	$\pm$	1.2	&	14.6	$\pm$	1.3	&	14.9	$\pm$	1.0	 & \textbf{15.7} $\pm$ 1.7\\
ROUGE-L	&	13.1	&	16.0	$\pm$	1.0	&	15.8	$\pm$	1.1	&	16.0	$\pm$	0.9	& \textbf{16.4} $\pm$ 1.5\\

\cmidrule[\heavyrulewidth]{1-6}
\end{tabular}
\end{adjustbox}
\caption{
The Pearson correlations with human judges when using  \imagineimage ($\textsc{IE}_{i}$) to augment ROUGE-1/2 and ROUGE-L on DUC2004. We compute four sets of \imagineimage similarity scores (mean$\pm$std) with dVAE, BigGAN, VQGAN, and StableDiffusion (SD). 
}
\label{tab:backbone_duc2004}
\end{table}

\paragraph{Sensitivity to Different Image Generation Backbones}
\label{sec:img_backbones}
In previous sections, we utilize StableDiffusion~\citep{Rombach2022HighResolutionIS} as the image generation backbone for \imagine. Here, we examine the influence of the image generation backbone on the evaluation performance of \imagine by conducting experiments on the DUC2004 dataset for summarization and comparing StableDiffusion with three alternative models: dVAE~\citep{dall-e}, BigGAN~\citep{Brock2019LargeSG}, and VQGAN~\citep{Esser2021TamingTF}. The results, as shown in~\cref{tab:backbone_duc2004}, indicate comparable performance of \imagineimage with dVAE and VQGAN, both of which outperform BigGAN across all metrics. StableDiffusion achieves the highest mean value, but also displays the largest variance among the models. These findings highlight the significance of considering the image generation architecture when evaluating text, as it can result in varying machine-generated images and affect the final evaluation outcomes.

\paragraph{Reliability of Machine-Generated Images}

\begin{table}[t!]
\begin{adjustbox}{width=\linewidth,center}
\begin{tabular}{ l | cccc}
\cmidrule[\heavyrulewidth]{1-5}
     & dVAE & BigGAN & VQGAN & StableDiffusion\\ \cmidrule[\heavyrulewidth]{1-5}
Entity Recall & 88.8\% & 41.2\% & 87.2\% &94.1\% \\
\cmidrule[\heavyrulewidth]{1-5}
\end{tabular}
\end{adjustbox}
\vspace{-2ex}
\caption{
Entity recall rate on the visualizations for Flickr30k captions. We report results for images generated by dVAE, BigGAN, VQGAN, and StableDiffusion.
}
\label{tab:entity_recall}
\end{table}

\begin{figure}[t!]
    \centering
    \includegraphics[width=\linewidth]{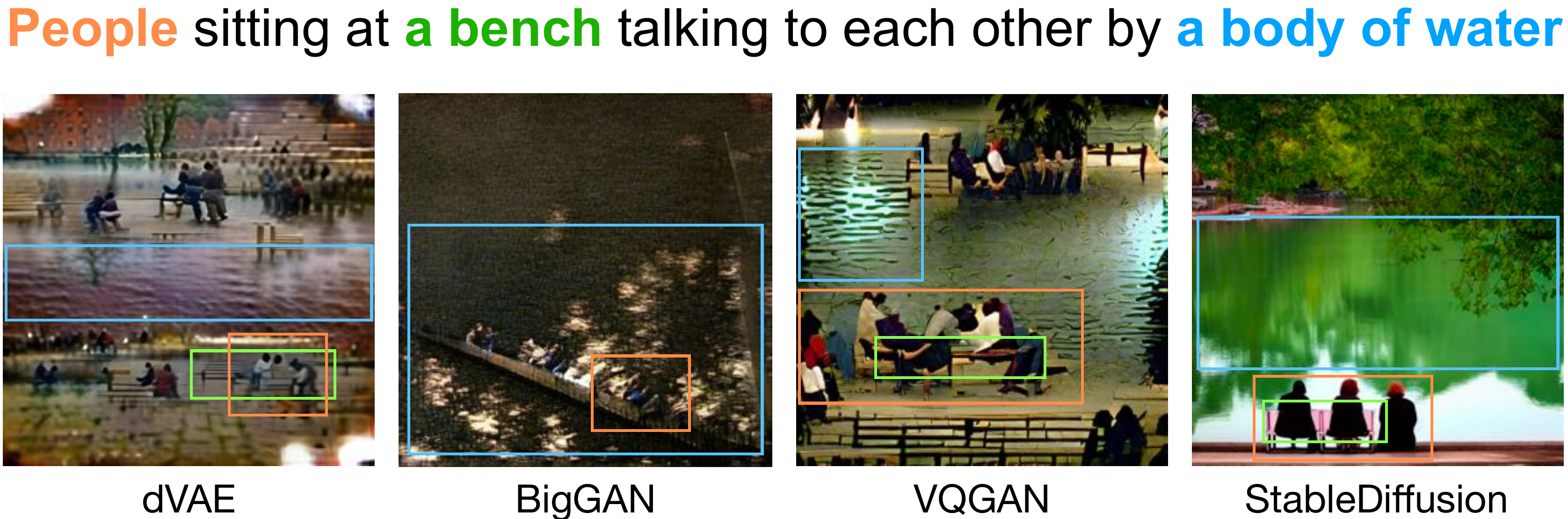}
\caption{
An example caption from Flickr30k Entities, and images rendered by dVAE, BigGAN, VQGAN and StableDiffusion. 
The bounding boxes point to the visualizations of the entities marked in the same color.
\label{fig:entity_recall_eg}
}
\end{figure}


The reliability of \imagine's visualization capability is further evaluated on the Flickr30k Entities dataset~\citep{Plummer2015Flickr30kEC}, which consists of annotated image captions. We randomly sample 100 captions and use the four generative backbones to render images. We present the captions and generated images to human annotators, and ask them to indicate if the entities mentioned in the captions are visually represented. The results, in terms of entity recall rates, are presented in~\cref{tab:entity_recall}. A higher recall rate indicates that the text-to-image generator is more capable of visualizing the content described in the text. The results show that StableDiffusion has the highest entity recall rate of approximately 94\%, followed closely by dVAE and VQGAN. In contrast, BigGAN has the lowest recall rate of around 41\%. An example of entity recall for a set of images generated by the four generative backbones is shown in~\cref{fig:entity_recall_eg}.

\begin{figure}[t!]
    \centering
    \includegraphics[width=\linewidth]{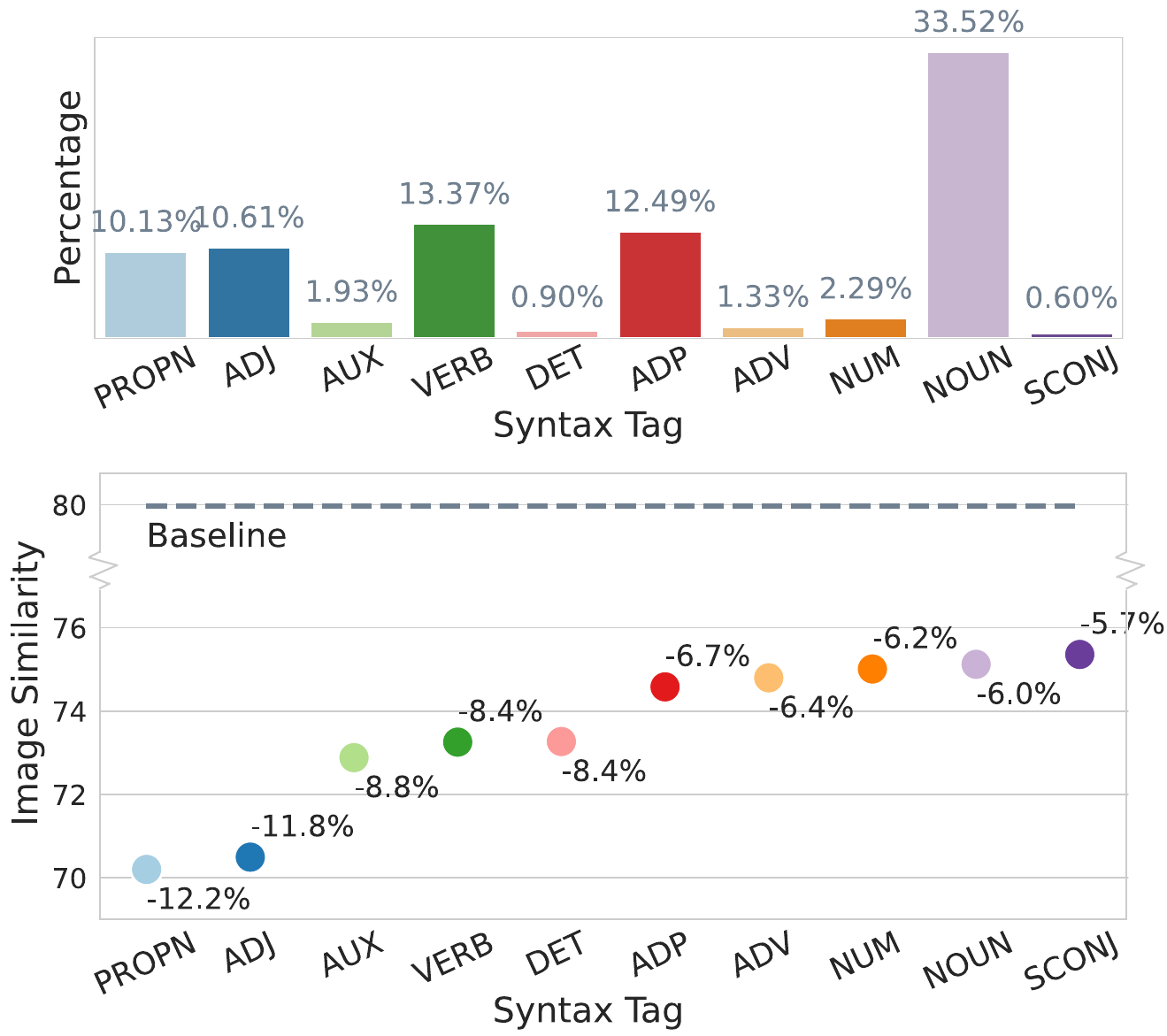}
\caption{
The influence on visualization when masking tokens of different syntax tags. 
Upper: The occurrence frequency of each syntax tag in DUC2004. 
Lower: The relative image similarity decrease after masking each syntax tag. Baseline: The average intra-group pairwise image similarity. The top-10 syntax tags that have the most significant impact on visualization are listed here.
\label{fig:pos_importance_duc2004}
}
\end{figure}

\paragraph{Syntax Importance to Machine-Generated Images}
\label{sec:syntax_importance}
We evaluate the significance of different syntax tokens in the image generation process using the DUC2004 summarization dataset. We utilized the Stanza~\citep{Qi2020StanzaAP} part-of-speech  (POS) tagger to parse the text and created ablated examples by masking out a token of a specific syntax tag.\footnote{We report Universal POS tags in this study: \href{https://universaldependencies.org/u/pos/}{https://universaldependencies.org/u/pos/}} The visual similarity of the images generated from the ablated examples is then compared to the visualization of the original text.
The results, as reported in Table~\ref{fig:pos_importance_duc2004}, indicated that the removal of PROPN and ADJ tags has a significant impact on the visualization results, resulting in a 12\% decrease in visual similarity. Conversely, removing NOUN tokens has a comparatively smaller effect. 
The most frequent NOUN, PROPN, and ADJ tokens in the DUC2004 dataset were listed in Table~\ref{tab:frequent_token_duc2004}. 
For DUC2004 built upon new clusters, PROPN and ADJ tokens cover concrete concepts such as nations, corporations, and celebrities, while NOUN tokens involve more abstract concepts such as government, party, and right. 
For this particular dataset, our \imagine approach pays more attention to PROPN and ADJ tokens that are easier to visualize by nature.
Further analysis for other dataset domains can be found in the Appendix.

\begin{table}[t!]
\begin{adjustbox}{width=\linewidth,center}
\begin{tabular}{ l | l }
\cmidrule[\heavyrulewidth]{1-2}
\textbf{POS Tag}     & \textbf{10 Most Frequent Tokens} \\ \cmidrule[\heavyrulewidth]{1-2}
NOUN    & \makecell[l]{president, minister, government, space, party, station,\\budget, game, right, arrest}  \\ \cmidrule{1-2}
PROPN   & \makecell[l]{U.S., Clinton, China, Korea, Gaza, Microsoft, Congo,\\Israel, Livingston, Lebanon}  \\ \cmidrule{1-2}
ADJ     & \makecell[l]{new, prime, Russian, international, Asian, possible,\\Cambodian, first, human, economic}  \\ 

\cmidrule[\heavyrulewidth]{1-2} 
\end{tabular}
\end{adjustbox}
\caption{
The most frequent NOUN, PROPN, and ADJ tokens in DUC2004.
}
\label{tab:frequent_token_duc2004}
\end{table}

\paragraph{Which \imagine Variant to Report?}
From ~\cref{tab:machine_translation_pearson,tab:summarization_pearson,tab:sentence_completion_pearson}, we can see a mixed trend of performance between the two \imagine variants. In general, \imaginetextandimage has smaller variances among repeated runs. 
Nevertheless, we would still suggest reporting both \imagine variants since they conduct comparisons from different aspects, with \imagineimage comparing similarity within the visual modality, while \imaginetextandimage compares cross-modal similarity.

\paragraph{\imagine as a Standalone Metric} 

Table~\ref{tab:imagine_standalone_pearson} presents the Pearson correlation with human evaluations on each dataset when utilizing the two \imagine variants as standalone metrics. The results reveal that both \imagine variants demonstrate a lower correlation compared to other metrics as reported in ~\cref{tab:machine_translation_pearson,tab:summarization_pearson,tab:sentence_completion_pearson}. Additionally, the scores produced by \imagine are not determinate, given the probabilistic nature of text-to-image models that generate various images with different random seeds. Hence, \imagine may not be an optimal choice as a standalone metric. Nonetheless, it is important to emphasize the capability of \imagine in introducing multimodal aspects to traditional text-only metrics. In this study, integrating \imagine with text-only metrics leads to an improvement in the Pearson correlation with human evaluations. 
Future work may explore alternative methods of integrating multimodal information in text evaluation.

\begin{table}[t]\centering
\begin{adjustbox}{width=\linewidth,center}
\begin{tabular}{lrrrrrrr}\toprule
&\textbf{IWSLT'14} &\textbf{WMT'19} &\textbf{DUC2004} &\textbf{GigaWord} &\textbf{AN(w/ ref)} &\textbf{AN(w/o ref)} \\\midrule
$\textsc{IE}_{i}$ &   19.1$\pm$1.5 &    13.8$\pm$1.7 &   10.6$\pm$1.5 &  15.9$\pm$1.1 &     18.9$\pm$1.5 &    16.8$\pm$1.9 \\
$\textsc{IE}_{t\&i}$ &    18.0$\pm$1.5 &    12.9$\pm$1.8 &    9.6$\pm$1.6 &     15.3$\pm$1.1 &    18.4$\pm$1.6 &   18.2$\pm$1.8 \\
\bottomrule
\end{tabular}
\end{adjustbox}
\caption{The Pearson correlation between \imagine variants and human assessments on each dataset. Here we use \imagineimage ($\textsc{IE}_{i}$) and \imaginetextandimage ($\textsc{IE}_{t\&i}$) as two individual metrics. AN: ActivityNet, ``w/ ref'': reference-based, ``w/o ref'': reference-free.}
\label{tab:imagine_standalone_pearson}
\end{table}

\section{Conclusion}

We present \imagine, a novel automatic evaluation metric for NLG that is based on machine imagination. Our experiments on five datasets across three different NLG tasks demonstrate the potential of incorporating \imagine similarity scores as a supplement to existing automatic NLG metrics, which can lead to improvement in their correlation with human evaluations in various scenarios.
In the future, it is interesting to explore effective ways of visualizing abstract concepts, and how to incorporate machine imagination efficiently. 
We hope our work can contribute to the discussion and advancement of multi-modal studies.


\subsection*{Limitations}
The current limitations of \textsc{ImaginE} include the length constraint of the CLIP text encoder, which is limited to 77 BPE tokens (including [BOS] and [EOS]), thus limiting its applicability to longer text generation tasks such as story generation or document summarization.
As a metric that relies on ``machine imagination'', \textsc{ImaginE} is limited by the inherent limitations of the generative models for images. The non-determined nature of machine-generated images can lead to non-determined \textsc{ImaginE} scores. Possible solutions to mitigate this issue includes but are not limited to fixing the random seeds or repeating the evaluation process several times to reduce the variance effect. Additionally, it remains a challenge for machines to properly visualize certain abstract concepts or numerical values, which could limit the scope of \textsc{ImaginE}'s applicability.

\subsubsection*{Ethical Statement}
Our study has received IRB exempt status and the estimated hourly wage paid to MTurk annotators is \$12. It is important to note that our ``imagination'' approach may raise questions of fairness if the training dataset for CLIP or StableDiffusion contains any biases. This could result in a tendency for \imagine to generate certain types of images based on what it has seen in the training data. While we did not observe such issues in our study, it is important to consider that such unfair behavior would undermine the effectiveness of \textsc{ImaginE} as an evaluation tool.

All of the datasets used in our study on machine translation, abstractive text summarization and open-ended text generation tasks are publicly available. We use the public repositories to implement \textsc{ImaginE}. The implementations of image generators used in our study are DALL-E(dVAE+CLIP),\footnote{\href{https://github.com/openai/DALL-E}{https://github.com/openai/DALL-E}} 
Big-Sleep(BigGAN+CLIP),\footnote{\href{https://github.com/lucidrains/big-sleep}{https://github.com/lucidrains/big-sleep}}  
VQGAN+CLIP,\footnote{\href{https://github.com/nerdyrodent/VQGAN-CLIP}{https://github.com/nerdyrodent/VQGAN-CLIP}} and StableDiffusion.\footnote{\href{https://huggingface.co/CompVis/stable-diffusion-v1-1}{https://huggingface.co/CompVis/stable-diffusion-v1-1}}

\subsection*{Acknowledgement}
The research was sponsored by the U.S. Army Research Office and was accomplished under Contract Number W911NF-19-D-0001 for the Institute for Collaborative Biotechnologies. This work was also supported by the National Science Foundation award \#2048122. The views and conclusions contained in this document are those of the authors and should not be interpreted as representing the official policies, either expressed or implied, of the U.S. Government. The U.S. Government is authorized to reproduce and distribute reprints for Government purposes notwithstanding any copyright notation herein.

\bibliography{anthology,custom}
\bibliographystyle{acl_natbib}

\appendix


\section{Appendix}

\subsection{Score Distributions}
\label{sec:rescaling}
\begin{figure}[htp]

\begin{subfigure}{\linewidth}
\includegraphics[width=\linewidth]{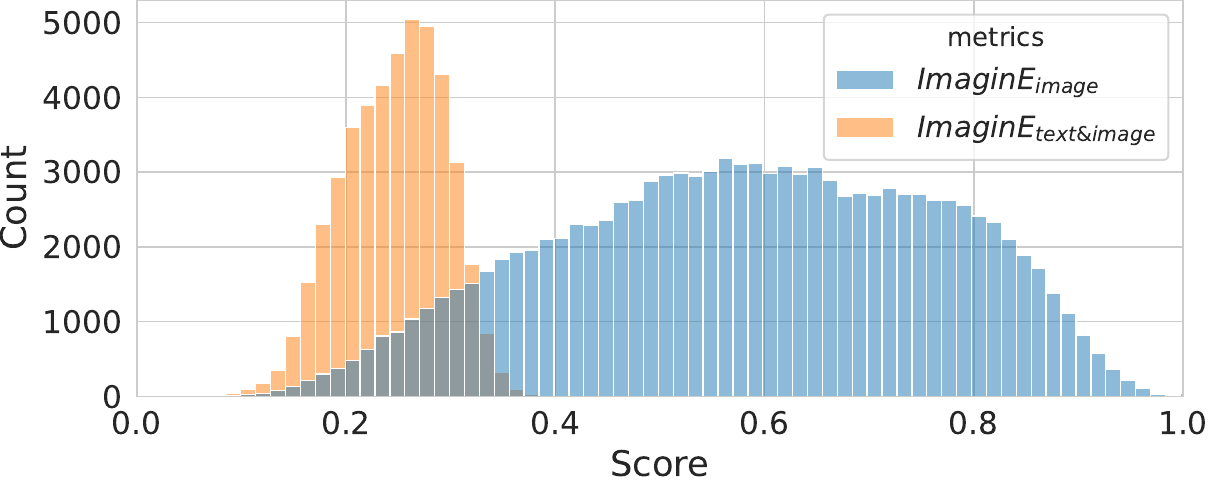}
\caption{Before re-scaling}
\end{subfigure}

\bigskip

\begin{subfigure}{\linewidth}
\includegraphics[width=\linewidth]{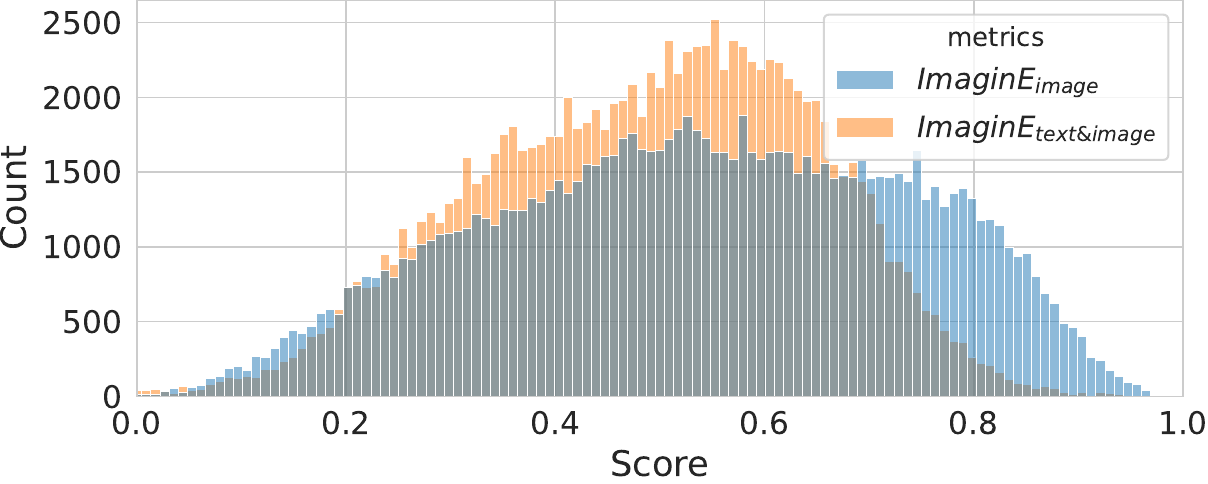}
\caption{After re-scaling}
\end{subfigure}

\caption{
The score distributions of \imagineimage and  \imaginetextandimage before and after re-scaling.
\label{fig:score_distribution}
}

\end{figure}

In this study, we use cosine similarity to evaluate the similarity between features, which yields a score distribution in the range of $[-1, 1]$. However, our results indicate that negative scores were not observed when computing the similarities between the features generated by CLIP. The score distributions of the two \imagine variants are depicted in Figure~\ref{fig:score_distribution}. Prior to re-scaling, the scores generated by \imagineimage typically fall within the range of $\left[ 0.1, 0.4\right]$, while those generated by \imaginetextandimage are within $\left[ 0.1, 1.0 \right]$. Following re-scaling, both \imagine metrics are linearly transformed to lie within the range $[0, 1]$.

\subsection{Syntax Importance to Imaginations}

\begin{figure}[htbp]
    \centering
    \includegraphics[width=\linewidth]{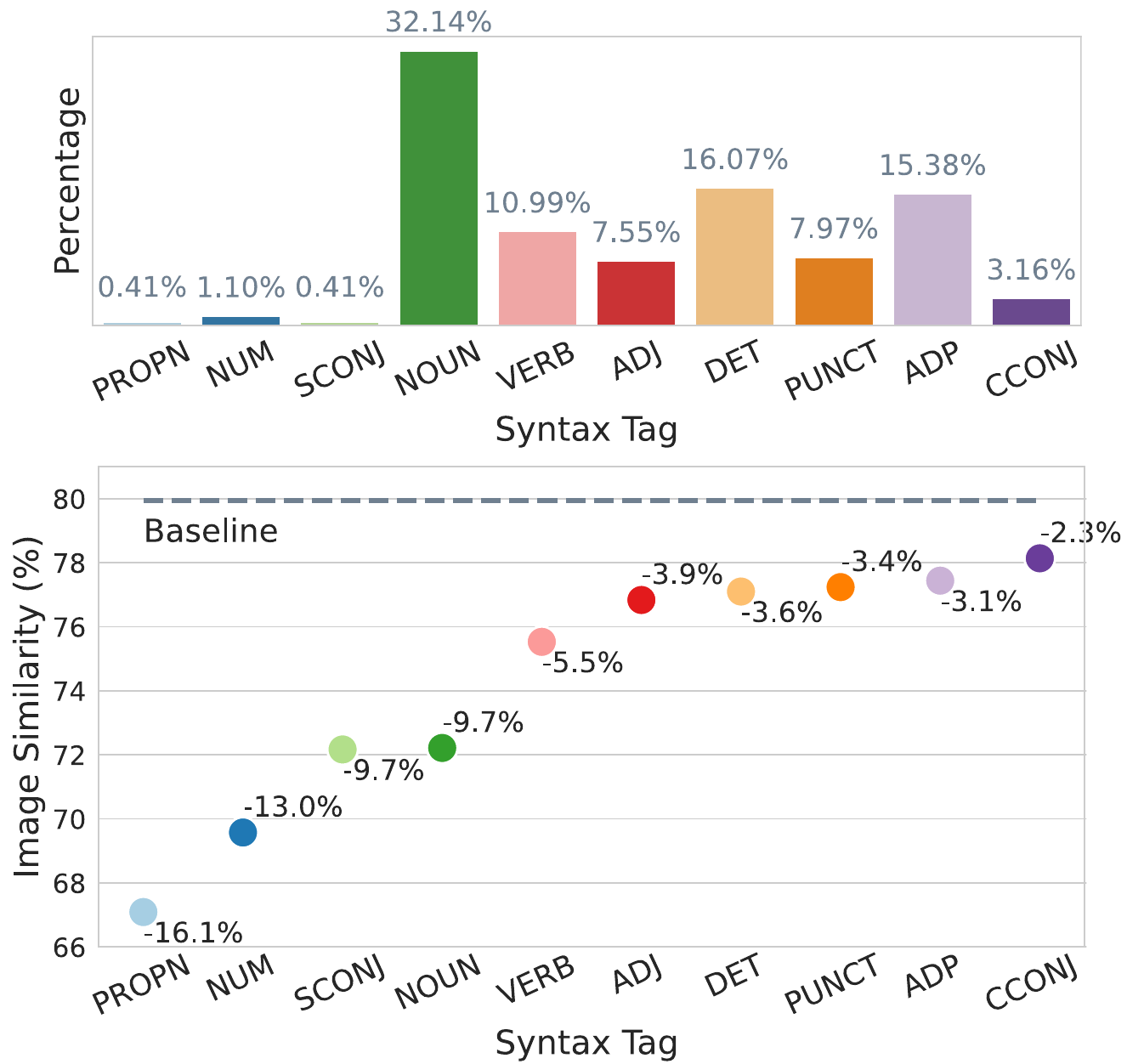}
\caption{
The influence on visualization when masking tokens of different syntax tags. 
Upper: The occurrence frequency of each syntax tag in Flickr30k. 
Lower: The relative image similarity decrease after masking each syntax tag. Baseline: The average intra-group pairwise image similarity. The top-10 syntax tags that have the most significant impact on visualization are listed here.
\label{fig:pos_importance_flickr30k}
}
\end{figure}

In \cref{sec:syntax_importance}, we discussed the impact of DUC2004 Part-of-Speech (POS) tags on the quality of generated images. In this section, we extend our examination to another dataset domain, the Flickr30k Entities dataset~\citep{Plummer2015Flickr30kEC}, which is an image captioning corpus.
While the domain of the Flickr30k Entities dataset is distinct from that of the DUC2004 (based on news articles), similar trends are observed. The results displayed in Figure~\ref{fig:pos_importance_flickr30k} also suggest that concrete concepts are easier to be visualized and play a more significant role in the visualization process, similar to the results observed in Figure~\ref{fig:pos_importance_duc2004}.

\end{document}